# Geometric Patterns of Meaning: A PHATE Manifold Analysis of Multi-lingual Embeddings

Wen G. Gong
wen.gong.research@gmail.com

## Abstract

We introduce a multi-level analysis framework for examining semantic geometry in multilingual embeddings, implemented through Semanscope (a visualization tool that applies PHATE manifold learning across four linguistic levels). Analysis of diverse datasets spanning sub-character components, alphabetic systems, semantic domains, and numerical concepts reveals systematic geometric patterns and critical limitations in current embedding models. At the sub-character level, purely structural elements (Chinese radicals) exhibit geometric collapse, highlighting model failures to distinguish semantic from structural components. At the character level, different writing systems show distinct geometric signatures. At the word level, content words form clustering-branching patterns across 20 semantic domains in English, Chinese, and German. Arabic numbers organize through spiral trajectories rather than clustering, violating standard distributional semantics assumptions. These findings establish PHATE manifold learning as an essential analytic tool not only for studying geometric structure of meaning in embedding space, but also for validating the effectiveness of embedding models in capturing semantic relationships.

## 1 Introduction

Understanding how semantic knowledge organizes in computational representations remains fundamental to NLP, yet current evaluation methods primarily focus on downstream task performance rather than examining the underlying geometric structure of embedding spaces. While multilingual embedding models demonstrate strong cross-lingual transfer capabilities, we lack systematic analytical tools to validate their effectiveness in capturing semantic relationships across multiple linguistic levels.

Recent work has explored semantic geometry at the word level [2], but comprehensive analysis spanning the full hierarchy of linguistic representation remains unexplored. Do embedding models handle all levels of linguistic structure equally well? How do purely structural components differ geometrically from semantic components across different writing systems?

We address these questions by utilizing PHATE manifold learning as a systematic analytical and validation framework for multilingual embeddings. Our approach examines four distinct levels of linguistic representation: (1) sub-character level examining purely structural components, (2) character/alphabet level analyzing writing system differences, (3) word/semantic level investigating conceptual organization, and (4) numerical level exploring Arabic number geometry. This framework not only reveals geometric patterns in embedding space but also validates embedding model effectiveness in capturing semantic relationships, providing concrete evidence for model capabilities and limitations across diverse languages.

## 2 Related Work

Geometry of Meaning: Gärdenfors [1, 2] established theoretical foundations for geometric approaches to conceptual organization, demonstrating that semantic domains organize spatially with meaningful geometric relationships. However, his framework primarily addresses word-level semantic organization, leaving character structural elements, writing system differences, and Arabic number geometry unexplored. Our multi-level analysis extends beyond this conceptual framework to investigate levels of linguistic representation not covered by existing geometric theories.

Manifold Learning in NLP: PHATE [3] has proven effective for biological data analysis by preserving both local neighborhood relationships and global transitional structures. While recent applications to text analysis show promise for document clustering [5], PHATE's potential as a systematic analytical and validation tool for embedding model effectiveness across multiple linguistic levels and typologically diverse languages remains unexplored.

| | ENU | CHN | DEU |
|---|---|---|---|
| Family | Germanic | Sino-Tib. | Germanic |
| Writing | Alphabet. | Logograph. | Alphabet. |
| Morphology | Analytic | Isolating | Synthetic |
| Core words | 278 | 265 | 265 |
| Networks | 62 | 123 | 90 |
| Numbers | 92 | 20 | 65 |
| Emoji | 50 | 100 | - |
| Total | 482 | 508 | 420 |

Table 1: Dataset Composition

Multilingual Embeddings: Models like Sentence-BERT Multilingual [4] create shared embedding spaces that demonstrate strong cross-lingual transfer performance, but the internal geometric organization of these representations lacks comprehensive empirical investigation. Understanding how these models handle different levels of linguistic structure—from purely structural components to complex semantic networks—is essential for improving multilingual architectures.

# 3 Methodology

## 3.1 Datasets

We constructed semantic datasets totaling 1,260 words across three typologically diverse languages as summarized in Table 1.

Core words cover parallel semantic domains: family/kinship, body parts, actions/verbs, emotions, nature/elements, animals, food, time/temporal, spatial/directional, abstract qualities, plus function words.

Morphological word networks capture language-specific word formation patterns: English (work/light families), Chinese (子-network: 孔子, 老子, 电子, 原子, 桌子, 杯子, etc.), German (haus/arbeit compounds).

Numbers include Arabic numerals (basic digits, magnitude progressions) and mathematical concepts. Emoji datasets compare modern digital pictographs with ancient pictographic writing.

## 3.2 Dimensionality Reduction Methods

After comparing 12 methods (see Appendix Table 2), we selected PHATE (Potential of Heat-diffusion for Affinity Transition Embedding). The PHATE algorithm applies a heat kernel to compute affinities between data points, then uses diffusion distances to preserve semantic relationships during dimensionality reduction (parameters: k=10 neighbors, alpha=10 decay, t=20 diffusion time). Unlike traditional methods that optimize for either local clustering (t-SNE) or hierarchical relationships (TriMap), PHATE uniquely balances both objectives through a diffusion operator that models data transitions as a Markov process.

## 3.3 Embedding Models

After evaluating 12 embedding models (see Appendix Table 3), we selected Qwen3-0.6B based on its superior cross-script separation, SOTA performance on HuggingFace MTEB leaderboard, and excellent clustering-branching balance across all hierarchical levels.

Our evaluation leverages three complementary access venues: (1) HuggingFace - direct model downloads requiring GPU/VRAM, (2) OpenRouter.ai - API-based access without local GPU requirements, and (3) Ollama - local deployment framework. OpenRouter's recent embedding API support in 2025 democratizes access to SOTA models.

## 3.4 Software Tool

We developed Semanscope, an interactive Streamlit application supporting 12 embedding models across three access venues, 12 dimensionality reduction methods, and both 2D and 3D visualizations. We plan to release Semanscope as open-source software upon publication.

# 4 Results

We present studies at three linguistic levels—character/alphabet systems, word-level semantics, and numerical concepts here, with additional studies at sub-character, sentence levels in appendices. In all multi-pane figures, subfigure labels (a), (b), (c), etc. follow sequential reading order: left-to-right, top-to-bottom.

## 4.1 Character/Alphabet System Analysis

Figure 1 shows six writing systems with distinct geometric signatures. Western alphabets (English/German in blue) cluster tightly reflecting shared Latin phonetic principles. Chinese characters (red) disperse widely across semantic space—each character carries meaning rather than pure phonetics. Digits in purple. Korean Hangul (cyan) and Japanese (yellow and grey) clusters are close by, positioned between alphabetic and logographic systems. Korean's striking linear organization reflects its intentional featural alphabet design in 1443, while Japanese occupies intermediate position reflecting its hybrid Kanji-Hiragana system. This validates fundamental script-family differences: alphabetic (compact phonetic), logographic (expansive semantic), and other intermediate.

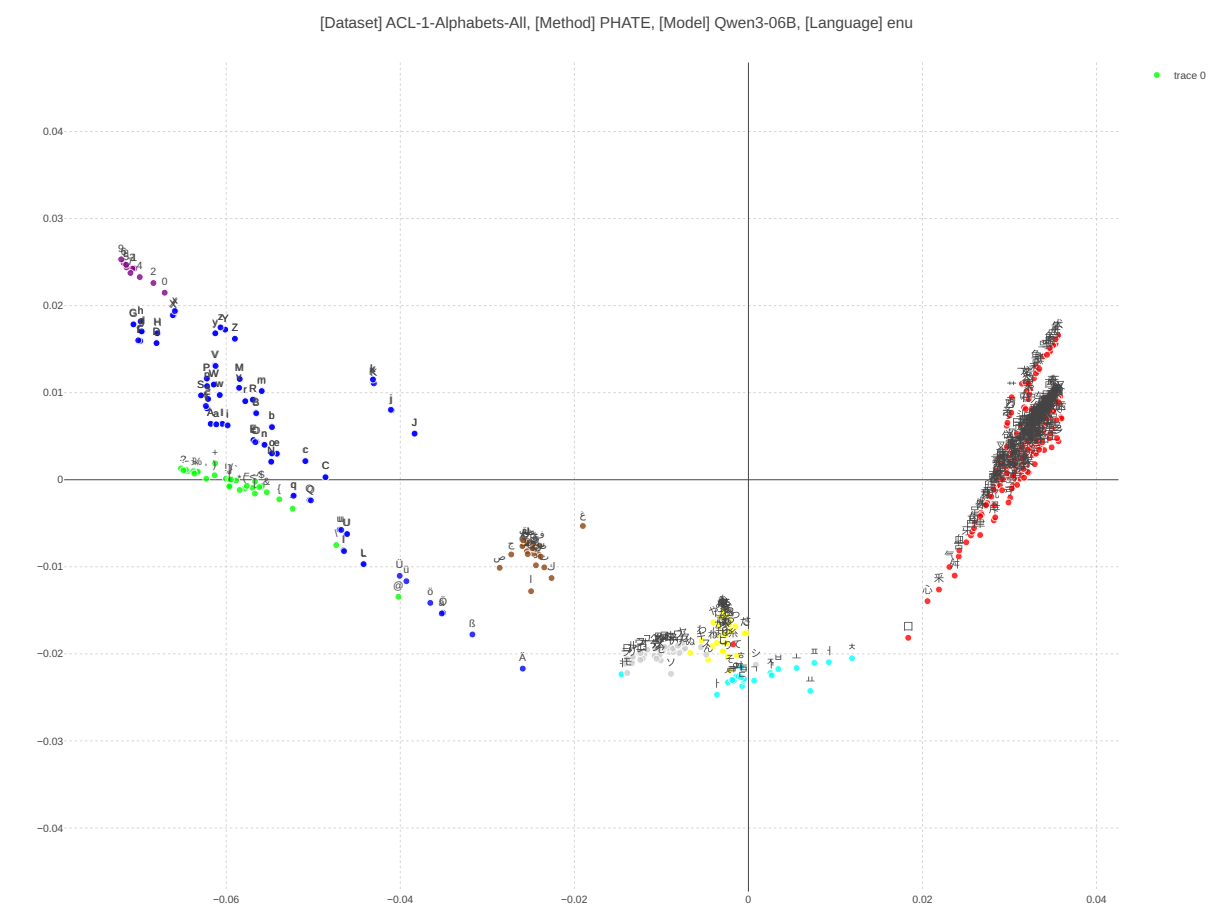

Figure 1: Six writing systems with distinct geometric signatures. Western alphabets (English/German in blue) cluster with much overlap. Chinese characters (red) disperse widely. Arabic (brown), Korean Hangul (cyan) and Japanese (yellow for hiragana, grey for katakana) clusters are close by. Digits in purple.

## 4.2 Word-Level Semantic Organization

Despite typological diversity—English (analytic), Chinese (isolating), German (synthetic)—Figure 2 reveals universal clustering-branching patterns. All three languages organize through two fundamental forces: (1) clustering - semantic similarity creates spatial proximity (family terms, animals, body parts form dense regions), and (2) branching - derivational relationships create directional extensions (work→worker→workplace). English shows work-family branching (upper-right) with animal/family clusters. Chinese exhibits similar patterns with family kinship clustering (upper-center) despite logographic writing. German demonstrates compound-specific organization while maintaining cluster-branch balance. This seems to validate universal cognitive organization principles across writing systems and morphological types.

While universal forces operate across languages (Figure 2), each exhibits distinctive morphological signatures (Figure 3). Chinese 子-character network: Balanced clustering-branching with philosophical terms (孔子/Confucius, 老子/Laozi) clustering separately from scientific terms (电子/electron, 原子/atom)—radial pattern reflects compositional flexibility. English work/light word network: Strong linear branching characteristic of derivational morphology (workplace, worker, working extend from "work")—affixes create systematic progressions. German haus/arbeit word network: Bilateral branching with Arbeit-compounds (left) and Haus-compounds (right) —two-hub structure demonstrates exceptional compound productivity. These patterns show language-specific morphological features: Chinese balances clustering-branching, English emphasizes branching, German amplifies bilateral extensions.

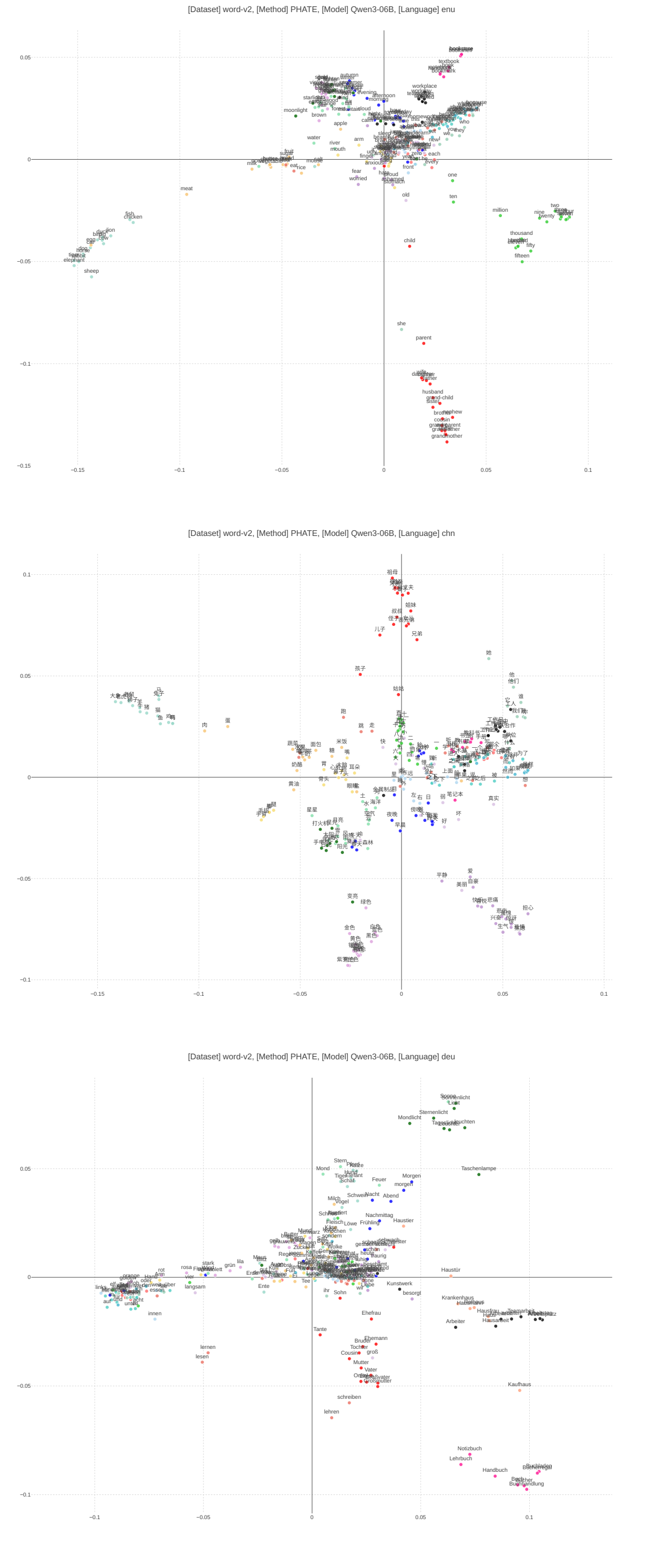

Figure 2: Universal clustering-branching patterns across (a) English, (b) Chinese, (c) German despite typological diversity.

Additional word-level analysis validating Gärdenfors' conceptual space framework is provided in Appendix A.2.

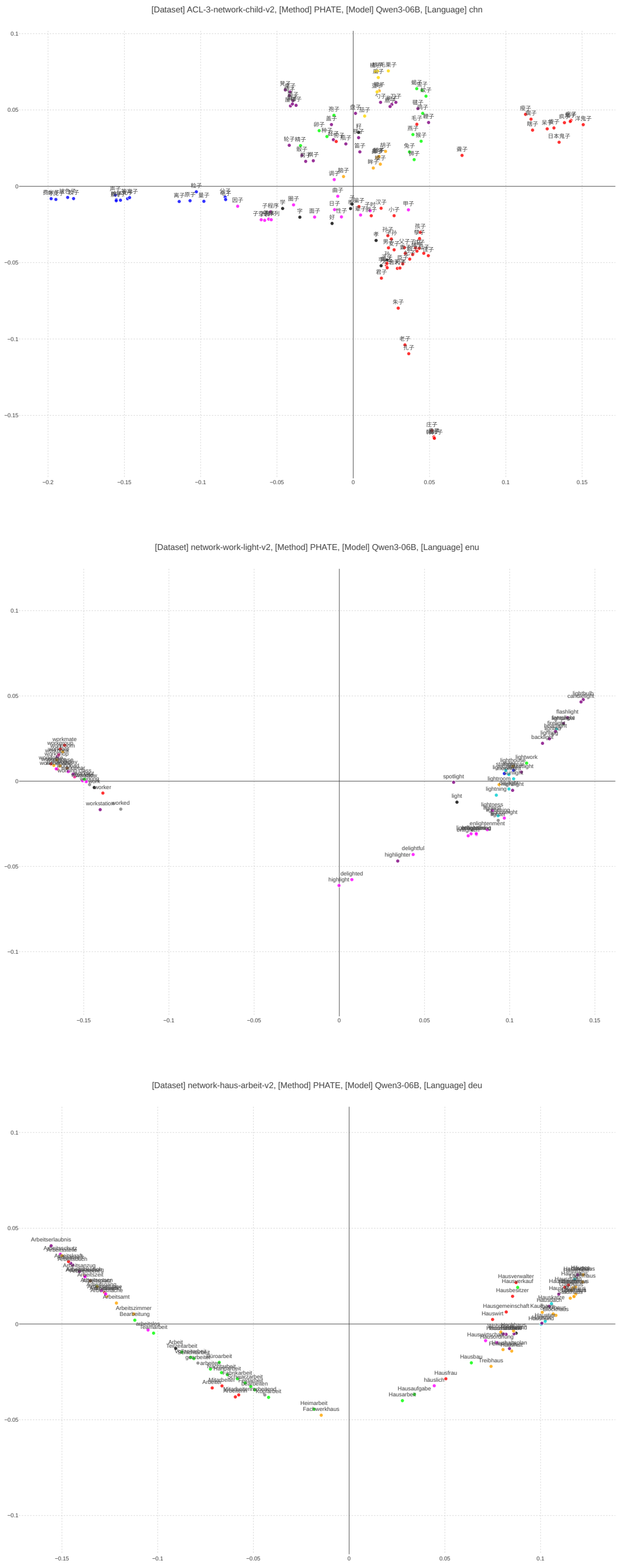

Figure 3: Language-specific morphological signatures: (a) Chinese 子-network shows balanced clustering-branching, (b) English work/light exhibits strong linear branching, (c) German haus/arbeit demonstrates bilateral branching.

## 4.3 Arabic Number Analysis

Figure 4 reveals critical geometric distinction: mathematical terminology (right: "subtraction", "computation", "multiplication") exhibits familiar clustering-branching patterns, while Arabic numerals display strong spiral pattern—a synthesis fundamentally different from pure clustering or branching. The powers of 10 (100000000→1000000→...→10→1) follow curved trajectories where magnitude progression (branching) combines with magnitude-order grouping (clustering) to create synthetic spirals. This demonstrates three emergent patterns across linguistic levels: (1) Pure clustering - high similarity, no directionality (family/animal terms), (2) Pure branching - clear directionality (work→worker→workplace), and (3) Spirals - simultaneous forces in ordinal domains (numerical magnitudes, temporal sequences, scale hierarchies). Spirals are not ad-hoc patterns but natural consequences of clustering-branching operating together in semantic spaces.

Critically, every combination of digits (0-9) forms a valid number with unique meaning, unlike alphabets or radicals where most combinations produce non-words—this fundamental difference in compositional density may explain the distinctive spiral geometry of numerical systems.

# 5 Discussion

PHATE analysis reveals systematic geometric organization across linguistic hierarchies through two fundamental forces—clustering (similarity attraction) and branching (directional extension) —that combine differently by level: (1) Character/Alphabet: Script-specific signatures (logographic dispersion vs. alphabetic clustering in Figure 1), (2) Word/Semantic: Universal clustering-branching across English, Chinese, German (Figures 2-3), with Appendix A.2 validating Gärdenfors' framework (nouns as regions, verbs as vectors, adjectives as dimensions) and A.3 showing sentence-level thematic gravity in poetry, (3) Arabic Numbers: Spiral synthesis where clustering-branching operate simultaneously (Figure 4), and (4) Sub-Character: Geometric collapse of purely structural radicals (see Appendix A.1).

Current embedding models exhibit selective proficiency: excellent at semantic relationships but failing at structural-semantic distinction (radical collapse) and showing variable writing-system competence. Appendix A.4 demonstrates this via emoji—Sentence-BERT complete failure, Gemini/Qwen3-8B strong integration. Appendices A.5-A.6 validate PHATE optimality among 12 methods and pattern robustness across 12 embedding models.

This work extends Gärdenfors' [2] conceptual spaces theory, establishing PHATE as both linguistic discovery tool and model diagnostic tool. Geometric consistency across typologically diverse languages suggests cognitive universals rather than artifacts.

## 6 Limitations

Our findings are subject to three key limitations: (1) while we validated patterns across 12 embedding models (see Appendix A.6, Table 3), geometric details vary by architecture, and our primary analysis uses Qwen3-0.6B's specific representations—non-transformer approaches might reveal different organizational patterns. (2) 2D PHATE projections represent "shadows" of multi-dimensional space, necessarily losing some semantic structure despite preserving key topological features. We validated PHATE's superiority over 11 alternative dimensionality reduction methods (Appendix A.5, Table 2), confirming it uniquely balances local clustering and global branching patterns. Our Semanscope tool provides interactive 3D visualizations that somewhat overcome the limitations of static 2D representations in print publications, enabling dynamic exploration of the full geometric structure. (3) our analysis covers only two language families (Sino-Tibetan, Germanic) and each dataset contains ∼400 core vocabulary words excluding technical terms and polysemous senses—broader claims require validation across additional families (Romance, Semitic, Bantu, Austronesian) and substantially larger lexical coverage.

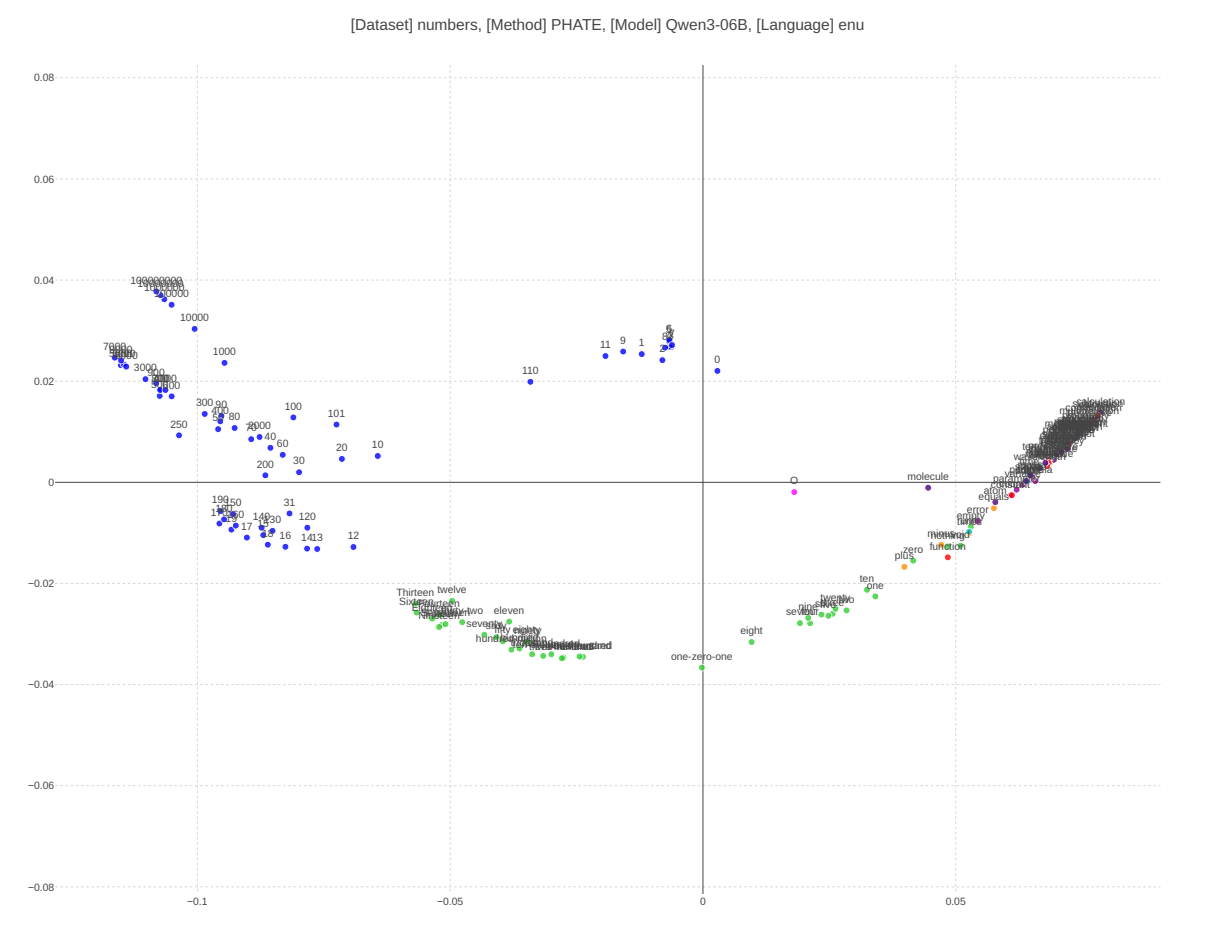

Figure 4: Arabic numerals display strong spiral patterns while mathematical terminology exhibits familiar clustering-branching patterns.

## 7 Conclusion

We establish PHATE manifold learning as a comprehensive discovery and validation framework examining semantic geometry across multiple linguistic levels with concrete evidence of model effectiveness and limitations.

Our specific contributions are: (1) Unified geometric framework - semantic organization emerges from two fundamental forces (clustering, branching); spirals represent simultaneous force synthesis in ordinal domains. (2) Critical limitations identified - sub-character geometric collapse (structural-semantic distinction failure), character-level writing system biases, and numerical spiral synthesis loss in failing models. (3) Model quality assessment - 6-language, 12-model analysis establishes geometric visualization as diagnostic tool; Qwen3 achieves superior cross-script understanding (MTEB #1) while EmbeddingGemma-300M exhibits catastrophic collapse. (4) Universal patterns - clustering-branching confirmed across typologically diverse languages (English, Chinese, German), extending Gärdenfors' [2] framework to computational embeddings. (5) Semanscope tool - PHATE-based visualization framework planned for open-source release.

Furthermore, we establish PHATE as essential for multilingual embedding validation, moving beyond task-specific performance to provide direct geometric evidence of representational quality. Cross-model validation demonstrates robustness across training objectives and parameter scales. Our findings suggest next-generation architectures require specialized components for different linguistic levels rather than assuming single embedding spaces optimally represent all knowledge types.

## Acknowledgments

The author acknowledges AI assistance from Claude and Gemini for literature research, coding, data analysis, and manuscript preparation. The author is grateful to Dr. Zheng-quan Tan for suggesting manifold learning approaches to Chinese character analysis, and to Albert W. Gong for recommending Qwen3 embedding models.

## Appendix

### A.1 Sub-Character Structural Component Analysis

PHATE analysis reveals dramatic geometric collapse in Chinese structural radicals using Sentence-BERT multilingual embeddings, as shown in Figure 5. Purely structural components (lacking independent semantic meaning) collapsed into 2 dots, forcing semantic radicals into an artificial 2D plane. After removing purely structural radicals, natural 3D manifold structures emerge with clear clustering-branching patterns, demonstrating model-specific limitations in structural-semantic distinction for logographic writing systems.

### A.2 Analysis of Gärdenfors' Conceptual Space

Gärdenfors [1, 2] proposed word classes occupy distinct geometries: nouns as regions, verbs as vectors, adjectives as dimensions. PHATE visualization empirically validates these distinctions, as shown in Figure 6. Nouns form localized clusters (animals, family/kinship) representing stable regions. Verbs exhibit distributed transitional patterns with directional trajectories connecting semantic domains. Adjectives organize along gradient axes (size: small→big, temperature: cold→hot, valence: negative→positive), validating dimensional structure. Combined visualization confirms geometrically distinct organizational patterns within shared embedding space.

### A.3 Compositional Analysis of Poetry

Beyond individual words, we examine whether embedding models capture compositional semantic structure at the sentence level, where meaning emerges from word combinations rather than isolated lexical items. Poetry provides an ideal test case as its meaning is often intentionally ambiguous—if our analytical framework can expose such subtle semantic structure, it demonstrates genuine sensitivity to compositional nuance.

Sentence-level PHATE analysis reveals thematic gravity in classical poetry, as shown in Figure 7. Li Bai's "Quiet Night Thought" centers around 月光/moonlight reflecting homesickness themes. Wang Wei's poems cluster near 孤独/solitude capturing mystic loneliness. Wordsworth's complex word distribution reflects nature contemplation. Frost's "The Road Not Taken" shows clear gravitational pull toward road/path vocabulary. Poetry organiza-tion demonstrates how complete compositional units (lines, stanzas, poems) exhibit geometric coherence around central semantic cores.

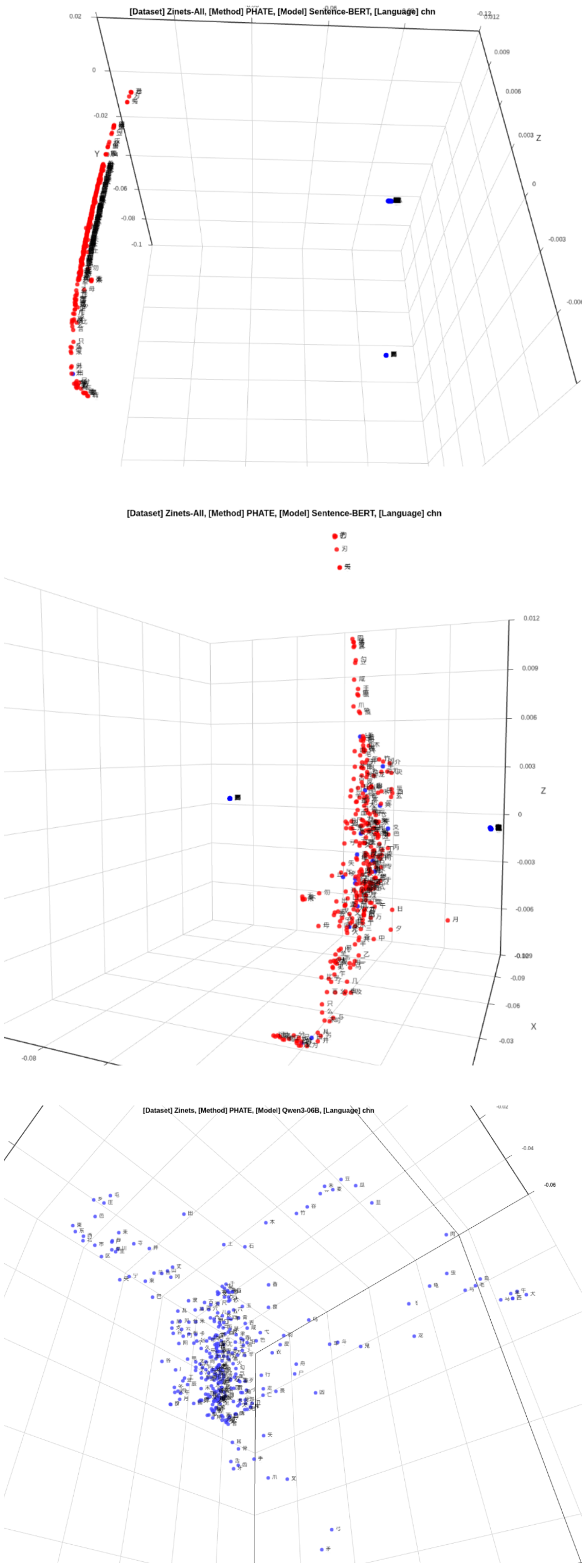

Figure 5: Chinese radical geometric analysis: (a) Dramatic collapse with structural radicals forcing 2D compression, (b) Alternative viewing angle showing central dense cluster, (c) Natural 3D organization (after removing purely structural radicals) revealing clustering-branching patterns.

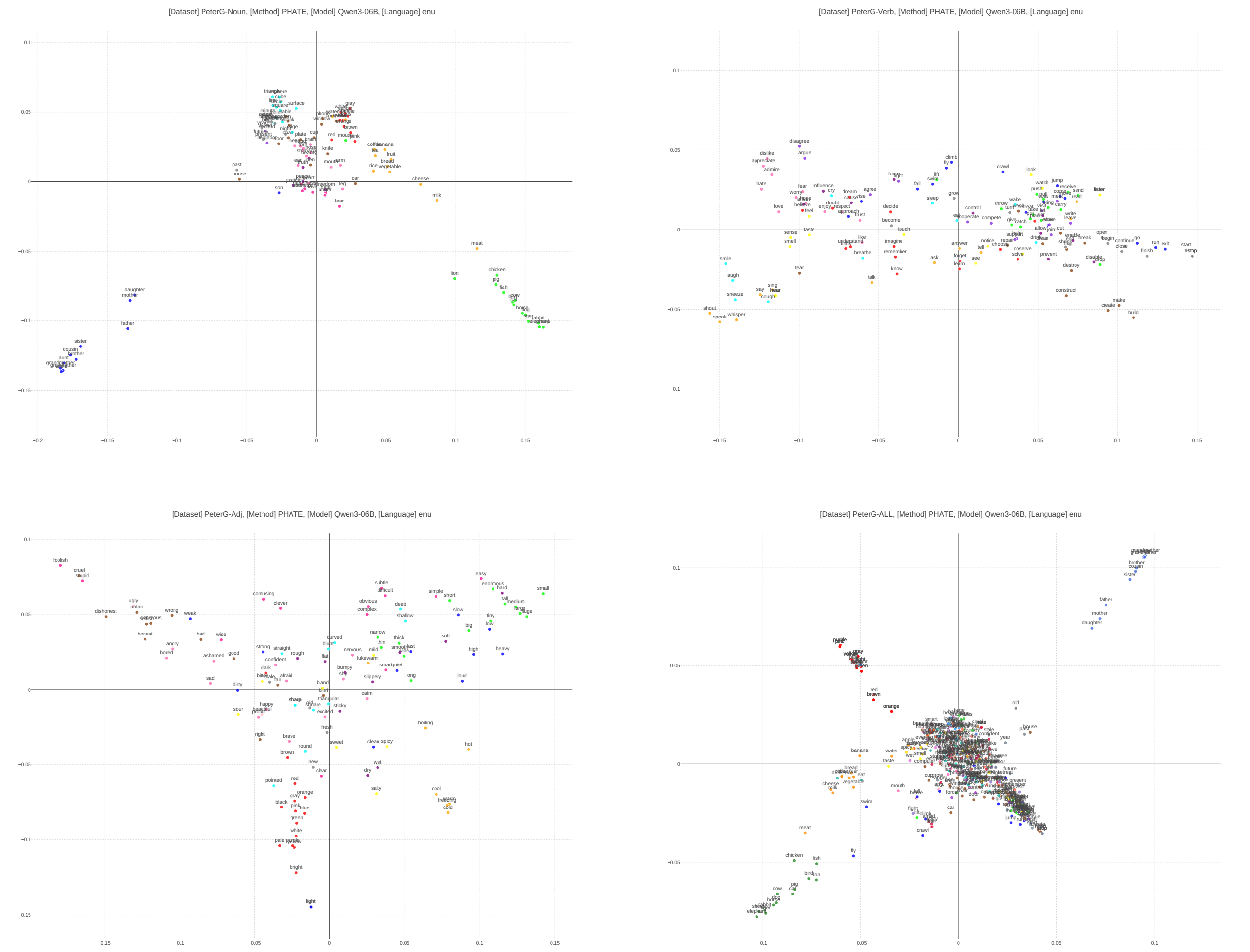

Figure 6: Validation of Gärdenfors' conceptual space framework: (a) Nouns form localized clusters (animals in green, family/kinship in blue), (b) Verbs exhibit distributed transitional patterns, (c) Adjectives organize along gradient dimensions, (d) Combined view shows geometrically distinct word-class patterns.

### A.4 Emoji Case-Study

Emoji represent a unique test of multimodal semantic understanding—they are visual pictographs that must align with corresponding textual meanings across languages. If embedding models truly capture semantic relationships rather than merely statistical co-occurrence patterns, emoji and their textual equivalents should occupy similar geometric regions in embedding space.

Cross-model evaluation of emoji understanding reveals striking capability gradients (Figure 8). Sentence-BERT exhibits complete emoji-text separation with emoji scattered linearly while text clusters compactly—demonstrating zero pictographic understanding. OpenAI text-3-small shows partial integration with reduced separation but distinct modality zones. Gemini-001 achieves strong emoji-text mixing with natural clustering-branching across both modalities. Qwen3-8B delivers exceptional integration where emoji and corresponding text occupy identical semantic regions ( fire adjacent to fire/火, water near water/水), validating superior multimodal understanding and suggesting specialized training on visual-semantic correspondences.

### A.5 Cross-Method Evaluation

Systematic comparison of 12 dimensionality reduction methods validates PHATE's superiority for semantic geometry analysis (Table 2 and Figure 9). t-SNE over-emphasizes local clustering, losing global branching structure. UMAP creates artificial separations between related semantic domains. TriMap focuses on hierarchical relationships at the expense of local clustering. PCA preserves global variance but obscures semantic organization. PHATE uniquely balances both objectives through diffusion-based distance computation, preserving clustering (family terms, animals) while maintaining branching trajectories (work→worker→workplace) and revealing spiral patterns (numerical magnitudes).

### A.6 Cross-Model Evaluation

Evaluation of 12 embedding models across parameter scales 300M-8B reveals parameter count $\neq$ geometric quality (Table 3 and Figure 10). Qwen3-0.6B (600M) achieves superior cross-script separation and clustering-branching balance despite smaller size compared to Qwen3-4B, demonstrating non-monotonic scaling. EmbeddingGemma-300M exhibits catastrophic geometric collapse with loss of all semantic structure—scheduled for detailed analysis in future work. Sentence-BERT shows selective proficiency: excellent word-level semantics but radical collapse and emoji failure. OpenAI text-embedding-3-large delivers strong performance across all hierarchical levels. Gemini-001 and Qwen3-8B represent current SOTA with exceptional multimodal integration and consistent geometric patterns.

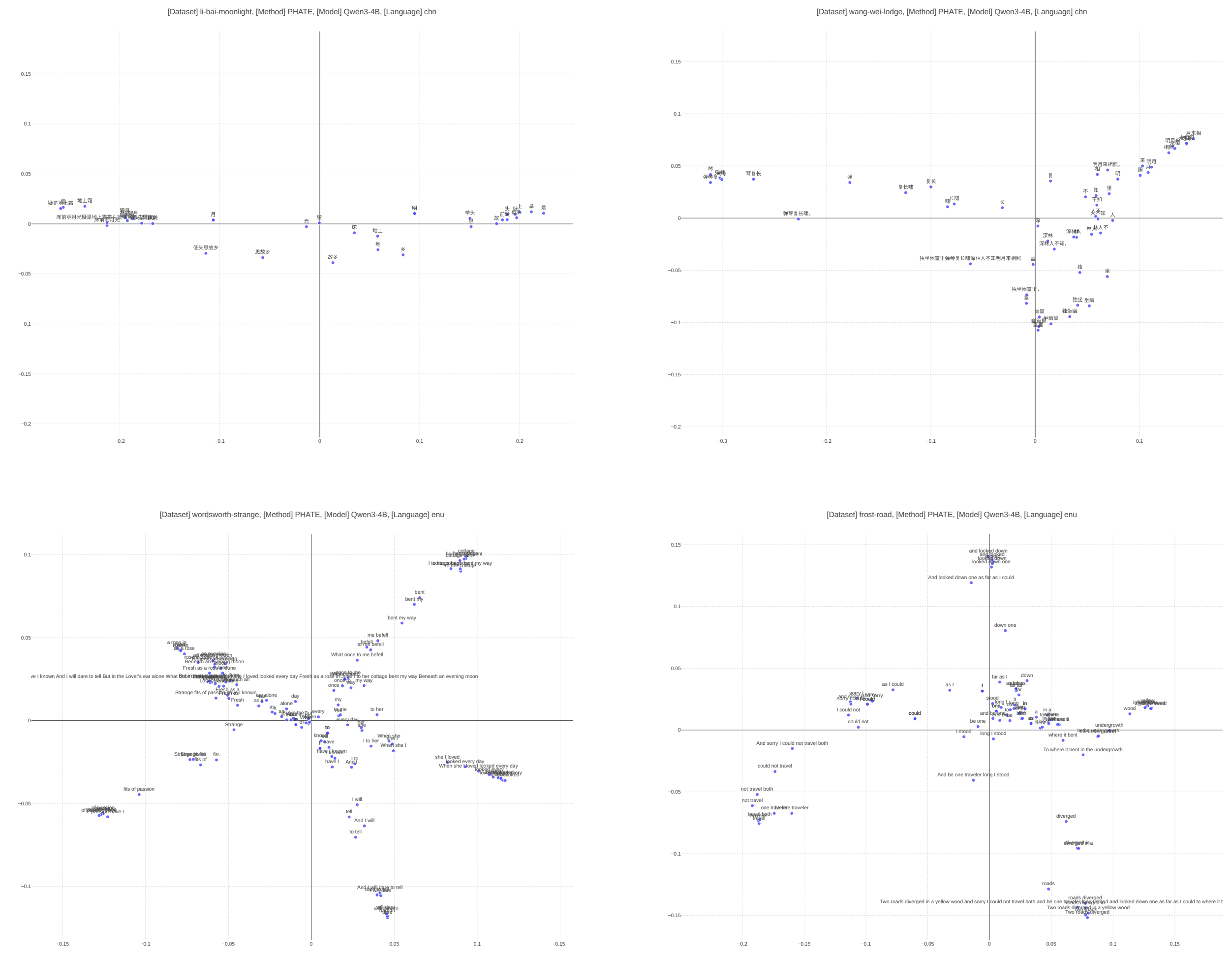

Figure 7: Thematic gravity in classical poetry: (a) Li Bai - poem centers around 月光/moonlight, (b) Wang Wei - clustering near 孤独/solitude reflecting mystic loneliness, (c) Wordsworth - nature contemplation patterns, (d) Frost - gravitational pull toward road/path vocabulary.

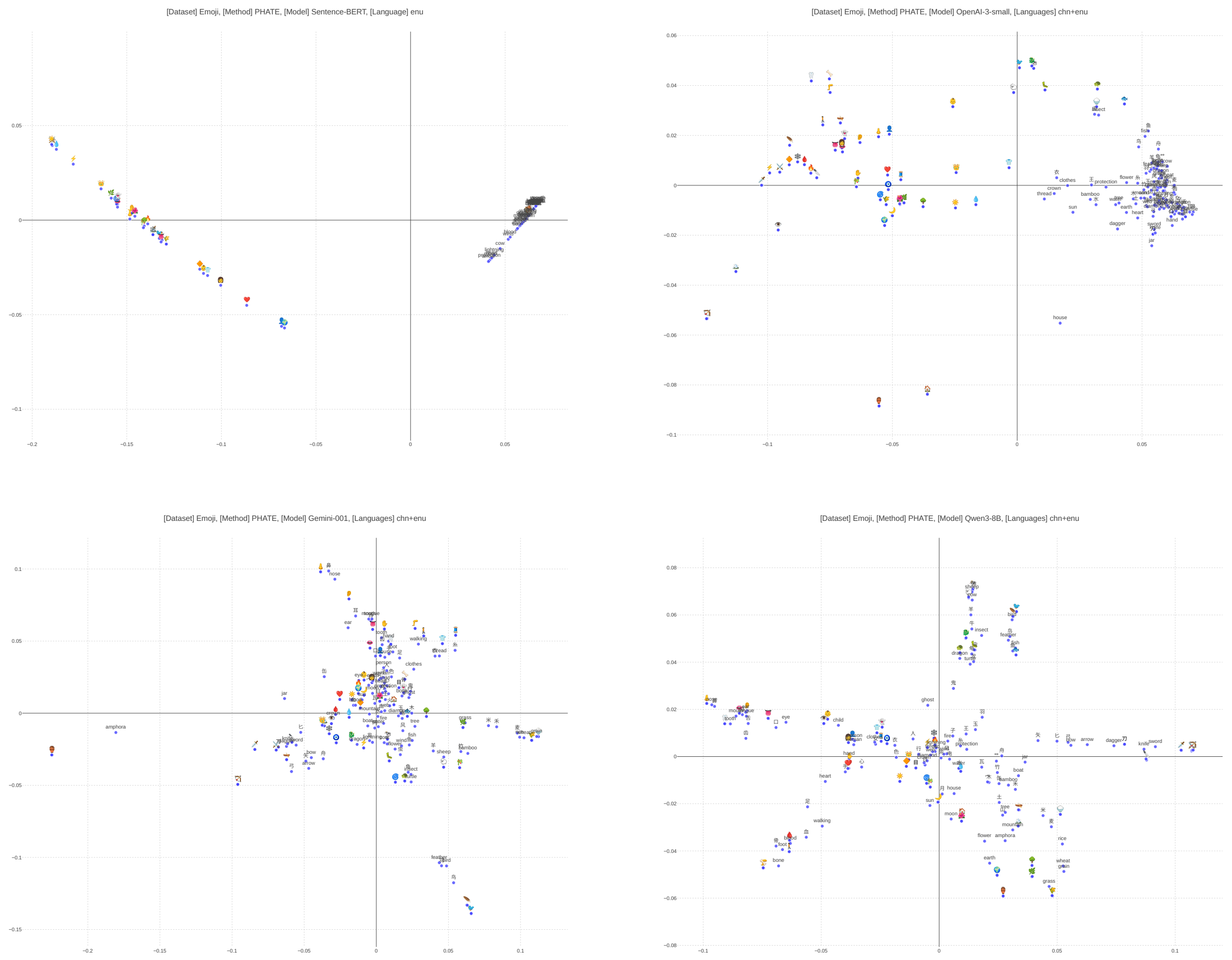

Figure 8: Emoji understanding capability gradient: (a) Sentence-BERT - complete emoji-text separation (failed), (b) OpenAI text-3-small - partial integration, (c) Gemini-001 - strong mixing across modalities, (d) Qwen3-8B - exceptional integration with emoji/text co-location.

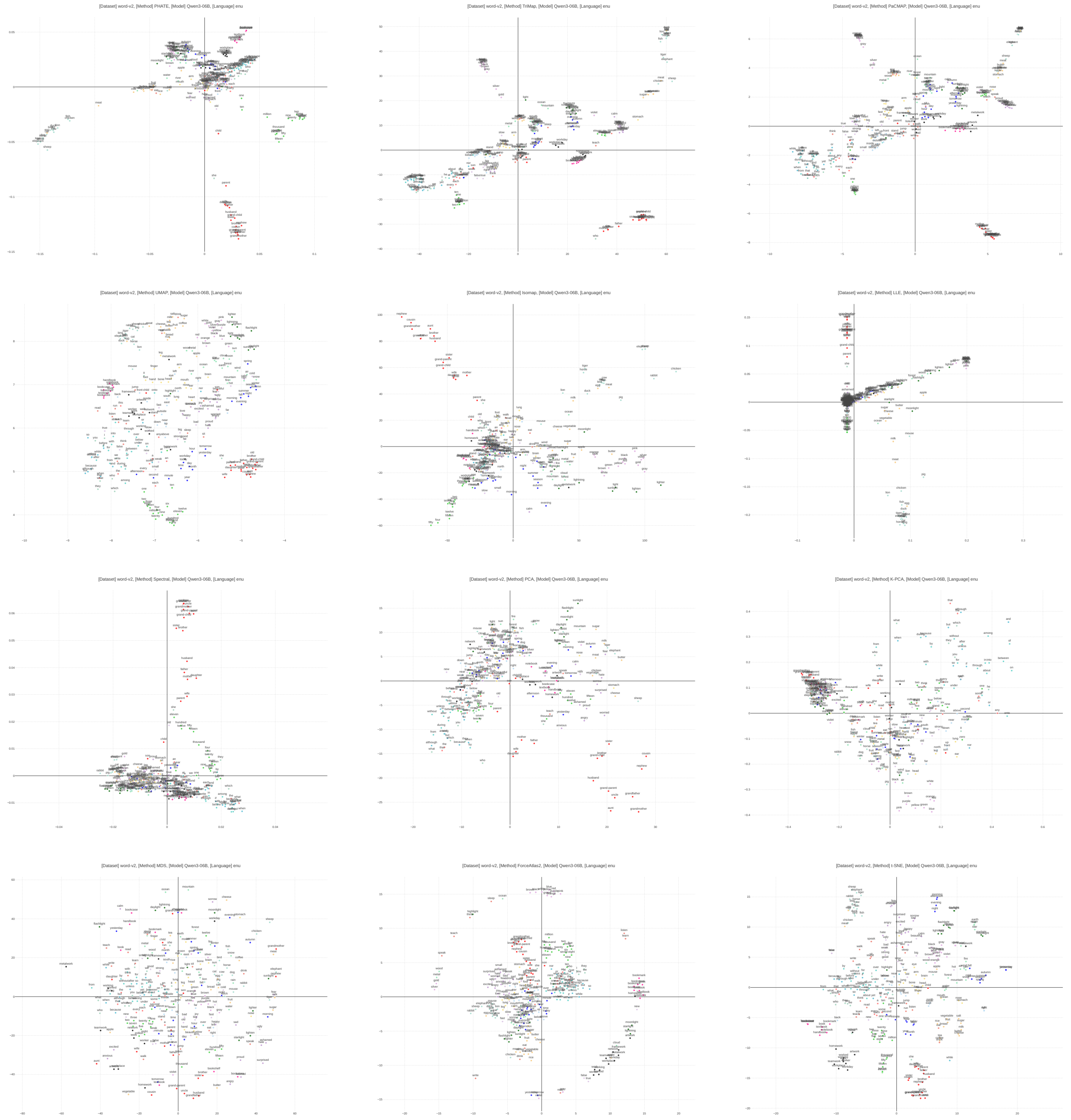

Figure 9: Cross-method comparison in a 4×3 grid: (a) PHATE, (b) TriMAP, (c) PacMAP, (d) UMAP, (e) IsoMAP, (f) LLE, (g) Spectral, (h) PCA, (i) Kernel PCA, (j) MDS, (k) ForceAtlas2, (l) t-SNE. PHATE uniquely balances local clustering and global branching patterns.

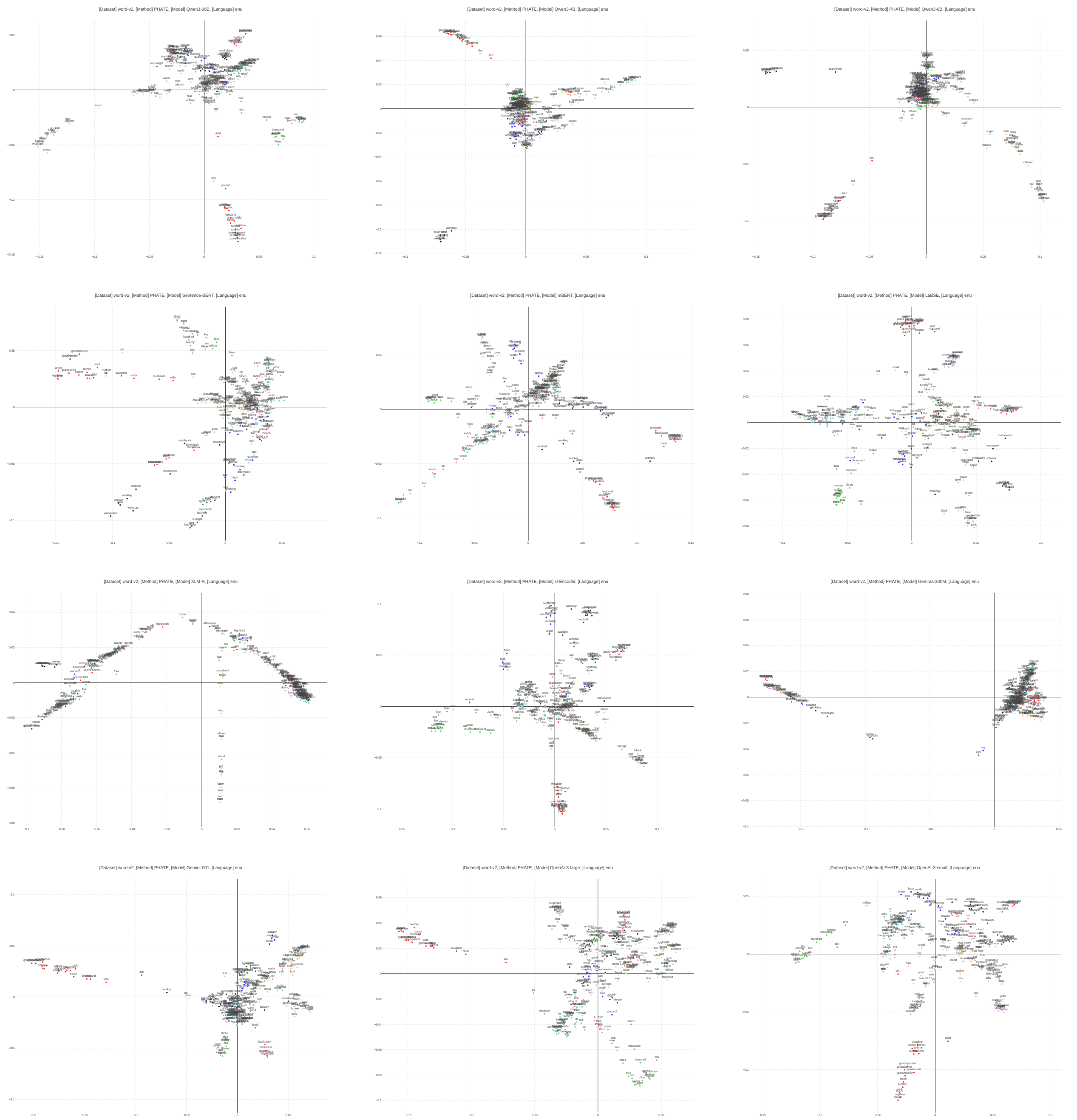

Figure 10: Cross-model comparison in a 4×3 grid: (a) Qwen3-0.6B, (b) Qwen3-4B, (c) Qwen3-8B, (d) Sentence-BERT, (e) mBERT, (f) LaBSE, (g) XLM-R, (h) Universal Encoder, (i) Gemma-300M (catastrophic collapse), (j) Gemini-001, (k) OpenAI-3-Large, (l) OpenAI-3-Small. Non-monotonic scaling evident: Qwen3-0.6B outperforms Qwen3-4B.

| Method | Type | Key Features | Chart Analysis |
|---|---|---|---|
| PHATE | Manifold | Optimal local/global balance | Clear semantic clustering with coherent branching patterns. Preserves both local neighborhoods and global morphological relationships. |
| TriMap | Manifold | Triplet-based learning | Second-best performance. Good clustering and some branching visible, but family terms scattered. |
| PaCMAP | Manifold | Balanced preservation | Shows distinct clusters but spatial arrangement doesn't reveal clear semantic relationships. |
| UMAP | Manifold | Topological preservation | Forms single massive central cluster with minimal internal structure. |
| Isomap | Manifold | Geodesic distance | Poor semantic organization. Most words compressed into central blob. |
| LLE | Manifold | Local linear embedding | Catastrophic failure. Most words compressed into horizontal line. |
| Spectral | Manifold | Graph Laplacian | Most semantic structure lost in central cluster compression. |
| PCA | Linear | Linear reduction | Spreads words uniformly. Cannot capture non-linear semantic relationships. |
| Kernel PCA | Kernel | Non-linear PCA | Uniform word distribution without clear clustering. |
| MDS | Distance | Multidimensional scaling | Better than linear methods but lacks clear clustering-branching distinction. |
| ForceAtlas2 | Force | Graph-based layout | Decent semantic clustering but lacks coherent branching patterns. |
| t-SNE | Probabilistic | Strong local clustering | Excellent tight clustering but sacrifices global structure. |

Table 2: Dimensionality Reduction Method Evaluation. PHATE uniquely balances local clustering and global branching preservation.

| Model | Arch | Focus | Params | Clustering | Branching |
|---|---|---|---|---|---|
| Qwen3-0.6B | Qwen | Multilingual | 600M | Excellent - Distributed clusters, clear domain separation | Excellent - Systematic morphological branching, balanced synthesis |
| Qwen3-4B | Qwen | Multilingual | 4B | Good - Multiple clusters, some central compression | Moderate - Present but less pronounced than 0.6B |
| Qwen3-8B | Qwen | Multilingual | 8B | Excellent - Strong distributed peripheral groups | Excellent - Multiple distinct trajectories, finest-grain structure |
| Sentence-BERT | BERT+Siamese | Sentence sim. | 278M | Excellent - Clear domain clusters | Excellent - Systematic work/family derivations |
| mBERT | BERT | MLM | 172M | Good - Multiple discrete clusters | Good - Clear work family branching |
| LaBSE | BERT+Trans | Lang-agnostic | 471M | Moderate - Scattered, less coherent | Moderate - Less systematic |
| XLM-R | RoBERTa | Cross-lingual | 270M | Good - Diagonal clustering pattern | Moderate - Some linear progressions |
| USE | Transformer | Multilingual | 110M | Good - Clear domains | Moderate - Less pronounced |
| Gemma-300M | Gemma | Multilingual | 308M | Failed - Catastrophic collapse | Failed - Complete absence |
| Gemini-001 | Gemini | Multilingual | N/A | Excellent - Well-distributed regions | Good - Systematic compositional patterns |
| OpenAI-3-Large | GPT | Text embed. | N/A | Excellent - Strong coherence, superior separation | Excellent - Clear derivational paths, well-preserved structure |
| OpenAI-3-Small | GPT | Text embed. | N/A | Good - Efficient multi-region clustering | Good - Adequate branching patterns |

Table 3: Embedding Model Architecture Comparison. Non-monotonic scaling evident: Qwen3-0.6B outperforms Qwen3-4B despite fewer parameters.